\documentclass[runningheads]{llncs}
\usepackage{graphicx}
\usepackage{placeins}
\usepackage{amsmath}
\usepackage{hyperref}
\usepackage{cleveref}
\usepackage{subfigure}
\usepackage[T1]{fontenc}

\begin{document}
	\title{Ammunition Component Classification Using Deep Learning}
	\author{Hadi Ghahremannezhad\inst{1} \and Chengjun Liu\inst{1} \and Hang Shi\inst{2}}
	\authorrunning{Ghahremannezhad, Liu, and Shi}
	\institute{New Jersey Institute of Technology\\
	Department of Computer Science\\
	Newark, NJ 07102 USA\\
	\and
	Innovative AI Technologies\\
	Newark, NJ 07201, USA\\}

\maketitle              

\begin{abstract}
Ammunition scrap inspection is an essential step in the process of recycling the ammunition metal scrap. Most ammunition is composed of a number of components, including case, primer, powder, and projectile. Ammo scrap containing energetics is considered to be potentially dangerous and should be separated before the recycling process. Manually inspecting each piece of scrap is tedious and time-consuming. We have gathered a dataset of ammunition components with the goal of applying artificial intelligence for classifying the safe and unsafe scrap pieces automatically. First, two training datasets are manually created from visual and x-ray images of ammo. Second, the x-ray dataset is augmented using the spatial transforms of histogram equalization, averaging, sharpening, power law, and Gaussian blurring in order to compensate for the lack of sufficient training data. Lastly, the representative YOLOv4 object detection method is applied to detect the ammo components and classify the scrap pieces into safe and unsafe classes, respectively. The trained models are tested against unseen data in order to evaluate the performance of the applied method. The experiments demonstrate the feasibility of ammo component detection and classification using deep learning.  The datasets and the pre-trained models are available at \href{https://github.com/hadi-ghnd/Scrap-Classification}{https://github.com/hadi-ghnd/Scrap-Classification}.


\end{abstract}

\section{Introduction} 

The non-usable ammunition goes through a rotary kiln incinerator (RKI) before recycling.
The safeness of the ammunition scrap should be confirmed before the process of recycling.
If the scrap still contains a considerable amount of energetics after the incineration process they are considered to be potentially dangerous.
Therefore, the scrap pieces are inspected in order to make sure there is no energetics left.
Manual inspection is a laborious, inaccurate, costly, and time-consuming step and is prone to human error.
Hence, there is a need for a reliable and effective method to inspect the ammunition scrap automatically.
Visual imaging and x-ray penetration are beneficial in detecting and discriminating the energetics remaining in the ammo scrap.

In this study, we have generated two datasets of visual and x-ray ammo images that are used for training a deep convolutional neural network (DCNN) to aid with the detection of explosive hazards on metallic ammunition scrap.
The goal is to sort the pure metal scrap from the scrap pieces that contain traces of explosive hazards.
The two classes are named MDAS (Material Documented as Safe) and MPPEH (Material Potentially Possessing Explosive Hazard), respectively.
Due to the lack of a sufficient number of x-ray images, several data augmentation techniques are applied as a pre-processing step.
The representative YOLOv4 object detection method \cite{bochkovskiy2020yolov4} is applied in order to train two DCNN models against the gathered training data.
The trained models are evaluated against the testing datasets according to appropriate measures to verify the effectiveness of the applied approach.

The remainder of this paper is organized as follows.
\Cref{sec_bg} the required background material about DCNNs and the YOLOv4 method is briefly described.
\Cref{sec_main} details the dataset generation and algorithms applied for training the deep CNN models that are used for ammunition component detection and explains the criteria used in the classification process.
In \cref{sec_expr} experiments conducted for evaluating the trained models are explained along with the assessment measures used to analyze the performance.
Finally, the paper is concluded in \cref{sec_con}.

\section{Background} \label{sec_bg}
Object detection and classification is one of the fundamental steps in many applications of computer vision and video analytics, such as robot vision \cite{o2018evaluating,guo2021real}, autonomous driving \cite{niranjan2021deep,feng2021review,feraco2022redundant}, and traffic monitoring \cite{liu2021smart,ghahremannezhad2020automatic,faruque2019vehicle,ghahremannezhad2020new,ghahremannezhad2021new,ghahremannezhad2020robust,shi2020statistical,ghahremannezhadreal,shi2021anomalous}.
Throughout the previous years many studies have been published that apply statistical methods, such as Support Vector Machine (SVM) \cite{osuna1997training}, efficient SVM (eSVM) \cite{chen2015eye}, Adaboost \cite{viola2001rapid}, the Bayesian Discriminating Features (BDF) method \cite{liu2003bayesian}, and discriminant analysis \cite{chen2014clustering}.

In recent years, deep convolutional neural networks have been popularly used for many tasks in computer vision, including object detection \cite{liu2020deep,zou2019object}.
A Convolutional Neural Network (CNN) is a deep learning algorithm that operates on an image as the input in order to train several parameters and extract high-level features for various tasks.
The architecture of Convolutional Neural Networks (CNNs) is well-suited to images in that it captures the spatial and temporal dependencies by applying appropriate filters.
The convolution layers in a CNN are usually coupled with activation functions and pooling layers for increasing the non-linearity and size reduction.

\begin{figure} [!t]
	\centering
	\includegraphics[width=0.8\linewidth]{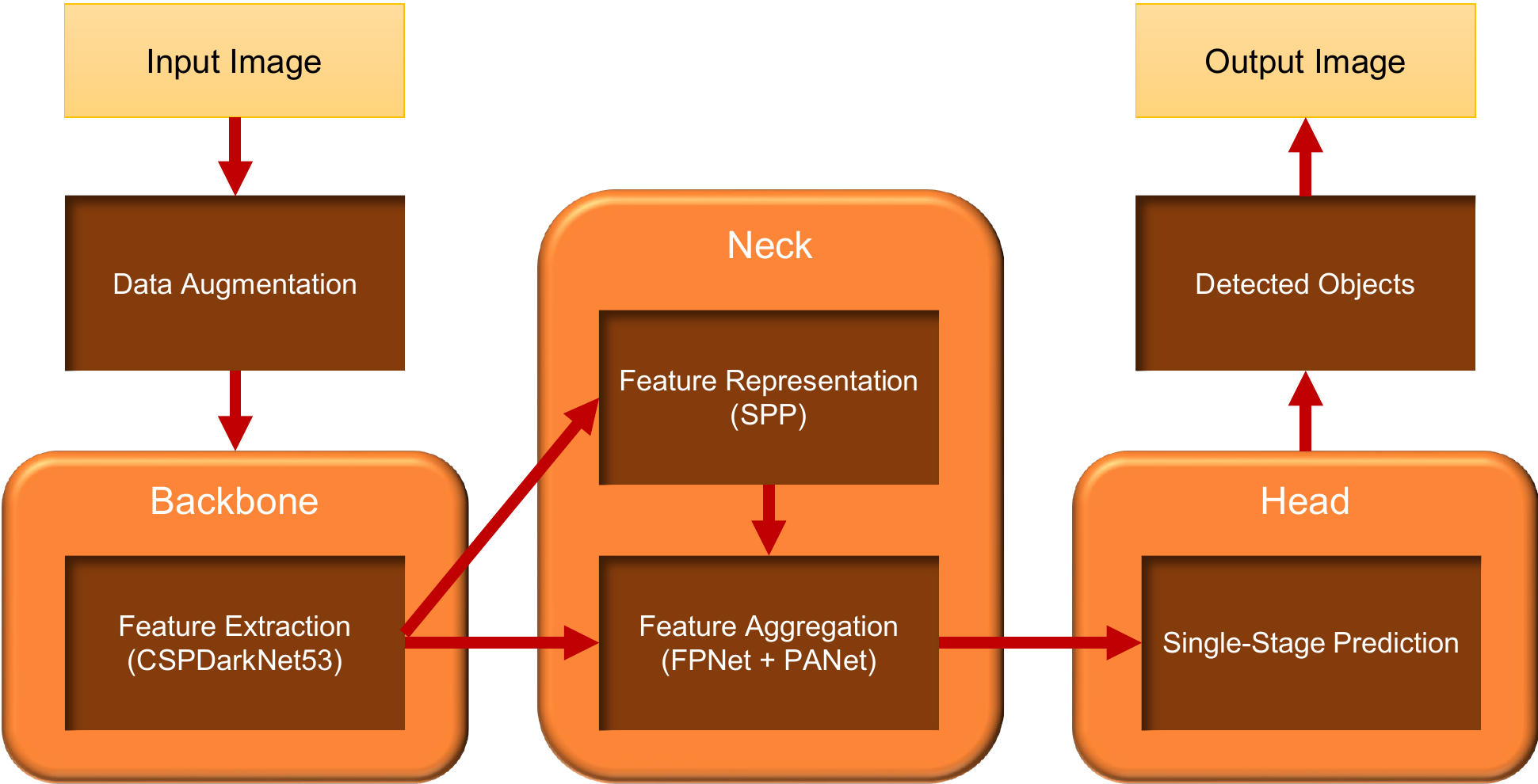}
	\caption{The main modules in the YOLOv4 model.}
	\label{fig:yolov4Model}
\end{figure}

One of the most representative object detection approaches in the You Only Look Once (YOLO) deep learning method was introduced in 2015 \cite{redmon2016you}. 
The core idea of this method is to divide the input image into an $S \times S$ grid where each cell of the grid either represents the background class or is used for the detection of the object the center of which falls in that cell.
At each cell, a predefined number ($B$) of bounding boxes are generated along with their corresponding confidence scores, which indicates how likely the box is to contain an object.
The probability of each object is multiplied with the intersection over union (IOU) of the predicted bounding box and the ground truth box to calculate the confidence scores.
As many object detection methods, YOLO utilizes the non-maximum suppression algorithm to remove the repetitive bounding boxes around each object and only keep the box with the highest score.

Since the development of YOLO at 2015, there have been several versions and varieties of this model proposed by introducing different improvements and alternations to the original model.
For instance, analyzing the error of the original YOLO approach showed a number of flaws including the production of a large number of localization errors and having a lower recall rate in comparison with the region proposal-based methods \cite{redmon2017yolo9000}.
In YOLOv2 \cite{redmon2017yolo9000} several techniques are employed in order to improve upon the original version.
These techniques include high-resolution classifier, batch normalization, direct location prediction, dimension clusters, and multi-scale training.
The Darknet-19 network was used as the classification backbone which consists of five max-pooling layers and 19 convolutional layers.

Later, YOLOv3 \cite{redmon2018yolov3} was introduced which further improved the robustness and efficiency of the previous versions.
In this version, the softmax layers are replaced with independent logistic classifiers and the binary cross-entropy loss is utilized in the classification process.
The Darknet-19 model is replaced with Darknet-53 and detections are performed in three different scales in order to deal with small objects, which was a problem for YOLOv2.
As opposed to the YOLOv2, which used the softmax function, YOLOv3 uses a multi-label classification approach and each bounding box can belong to several classes at the same time.

The latest official version of this method is YOLOv4 which is improves the performance of the previous version both in terms of mean average precision (mAP) and speed.
As seen in the \cref{fig:yolov4Model} the architecture of YOLOv4 consists of three distinct components, namely, the backbone, the neck, and the head.
The backbone network for feature extraction is the CSPDarknet53 which is used for splitting the current layer into two parts.
The first part passes through the convolution layers while the second part doesn't and the results are aggregated at the end.
The neck is the intermediate section between the backbone and the head and contains modified versions of the path aggregation network (PANet) and spatial attention module with the purpose of having a higher accuracy by information aggregation.
The head of the architecture represents the dense prediction block, which is used to locate the bounding boxes and final classification.
Similar to YOLOv3 the bounding box locations and object probabilities are calculated as the output of the model.

\begin{figure} [!t]
	\centering
	\includegraphics[width=0.8\linewidth]{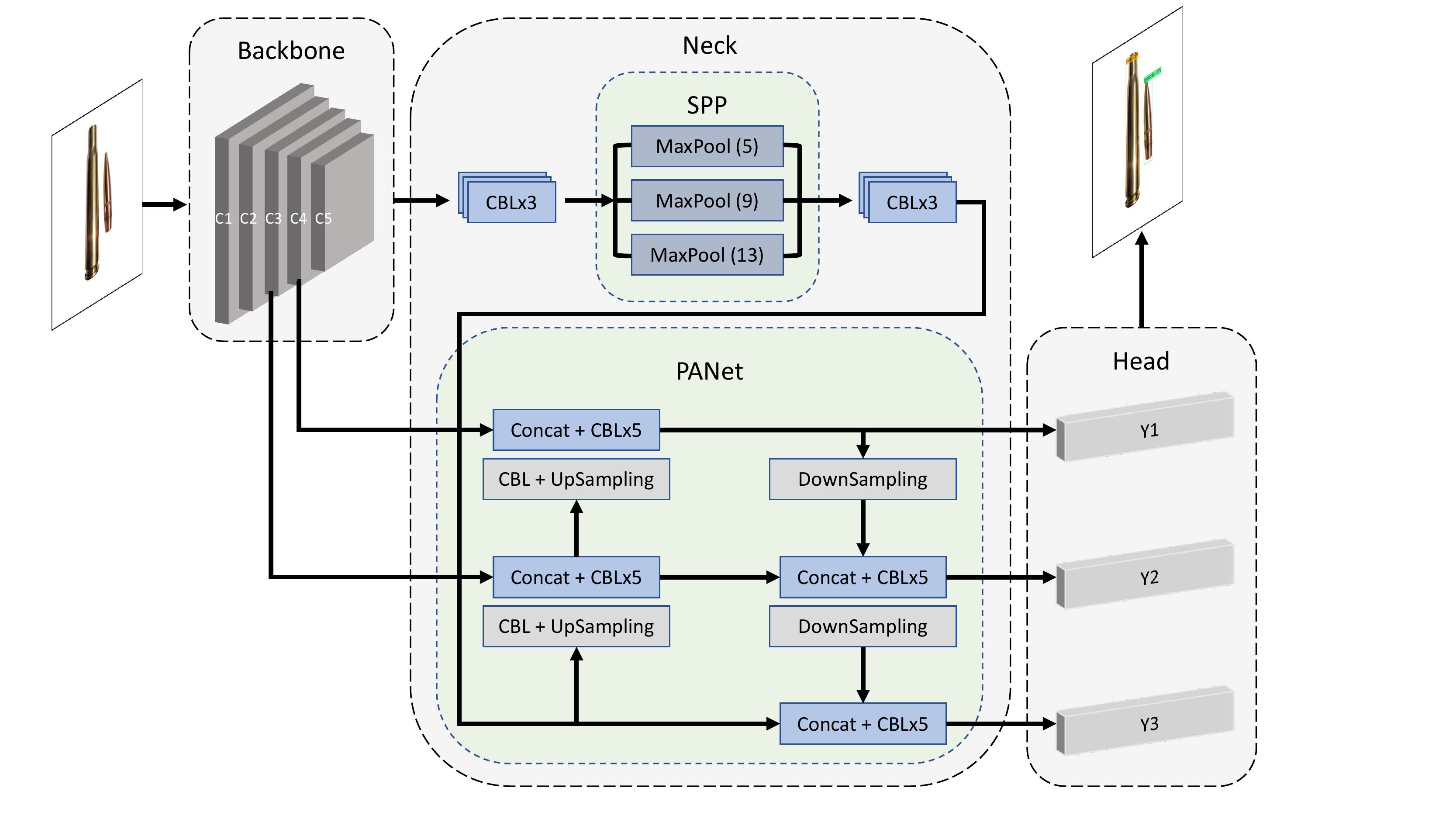}
	\caption{The system architecture of the YOLOv4 model.}
	\label{fig:yolov4Arch}
\end{figure}

Several additional sets of techniques are applied in YOLOv4 in order to further enhance the detection results, which are called bag of freebies and bag of specials.
The bag of freebies consists of various approaches, such as cut mix and mosaic data augmentation, drop block for regularization, self-adversarial training, and random training shapes.
On the other hand, the so-called bag of specials is a set of post-processing modules designed to considerably improve the accuracy with a slight increase in the inference time.
This set of techniques includes different modules including mish activation, cross-stage partial connections (CSP), the spatial pyramid pooling (SPP) block, the spatial attention module (SAM), path aggregation network (PANet), and the distance IoU non-maximum suppression.
\Cref{fig:yolov4Arch} illustrates the detailed architecture of YOLOv4 object detection method.

\section{Ammunition Component Detection and Classification Using Deep Learning} \label{sec_main}

We apply one of the most representative deep learning methods to date, the YOLOv4, for ammunition component classification in visual and x-ray ammo images. 
The training data sets are manually created using publicly available images. 
Specifically, two training data sets are created corresponding to the training samples from visual and x-ray images mostly captured from 50 calibers by vision cameras, transmission x-rays, or back-scatter x-rays.
Each image contains one or more ammo components or full-ammo. 
The goal is to detect each full-ammo or ammo component and classify it into one of the two classes, namely Material Documented as Safe (MDAS) and Material Potentially Possessing Energetic Hazard (MPPEH).
These classes indicate whether the scrap piece is safe or potentially hazardous.

\begin{figure} [!t]
	\centering
	\includegraphics[width=0.2\linewidth]{"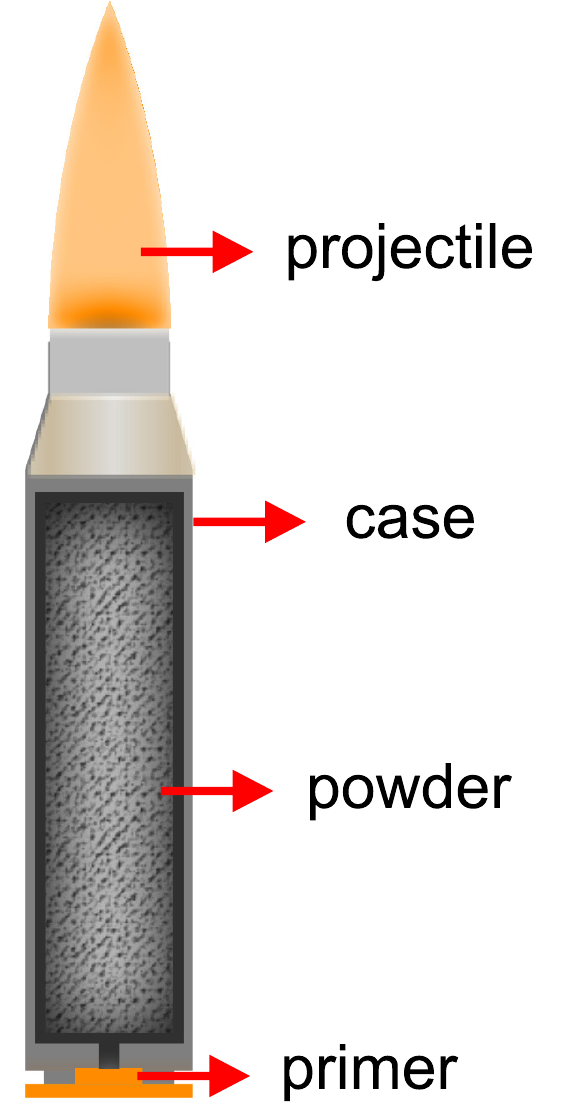"}
	\caption{Basic Components of Ammunition.}
	\label{fig:ammo}
\end{figure}

The basic components of ammunition are the case, gunpowder, primer, and projectile.
The case is a usually cylindrical container that holds the ammunition components together as one piece.
Various materials are used for the case, such as steel, copper, brass, plastic, and paper.
The powder is a chemical mixture and is converted to an expanding gas when ignited.
This component of ammunition is considered to be explosive hazard and should not remain in the scrap during the recycling process.
The primer is an explosive chemical compound that is used to ignite gunpowder when it is struck by a firing pin.
The projectile is the part of ammunition that is expelled from the barrel which is usually referred to as a bullet.
\Cref{fig:ammo} demonstrates the main components of ammunition.

Typically, ammunition scrap pieces are loose casings, loose projectiles, or full ammo with case and projectile in place, but missing the primer.
Blown out casings and burst projectiles are also likely to appear among the metallic scrap.
In an ideal scenario, there is no energetics left in the ammunition scrap and only the metal parts go through the recycling process.
However, in real-world situations, some of the ammunition is not destroyed well enough during the incineration and the powder remains inside the case.
This is considered to be potentially dangerous as there is an explosive hazard among the metallic scrap.
Therefore, there is a need for inspecting the scrap in order to make sure all the scrap pieces are safe to go through the recycling process.
Here, we have constructed a dataset of visual images and a dataset of x-ray images with the purpose of training a deep learning model for the detection and classification of safe and unsafe scrap pieces.

\subsection{Ammunition Component Detection in Visual Images}
We have gathered and annotated a sufficient number of visual images of ammunition components and full-ammo to be used as training and testing samples.
The YOLOv4 object detection method is applied in order to train three classes of scrap pieces.
Specifically, full-ammo, projectile, and case are the three classes that are used to train the deep CNN.
Since there is a good chance of a full-ammo to contain energetics in its case, we have considered this class to represent the unsafe samples, called Material Potentially Possessing Energetic Hazard (MPPEH).
On the other hand, the separated casings or projectiles are assumed to be safe with a high degree of certainty and are classified as Material Documented as Safe (MDAS).

In addition to the original YOLOv4 structure, we have also used a shallower lightweight network, called tiny-YOLOv4 as the backbone in order to increase the computational speed with a negligible drop in the accuracy \cite{montalbo2020computer}.
The lightweight structure contains 29 layers compared to the original one with more than a hundred.
The Cross Stage Partial (CSP) model is derived from the DenseNet architecture, which concatenates the previous input with the current input prior to reaching to the dense layer.
The backbone of the tiny-YOLOv4 includes an input layer followed by 18 convolutional layers, 9 routes, 3 max-pooling layers, and a detection layer based on YOLOv3 at the end.
Several features of each input image are extracted by the convolutional layers.
There are interchangeable $3\times3$ and $1\times1$ receptive filters striding over the input image to generate feature maps, which are passed through other layers of the network. 
The leaky-ReLU activation function is applied at the convolutional layers in order to increase the feature size.
Routes are designed to improve the gradient flow throughout the layers.

\subsection{Ammunition Component Detection in X-Ray Images}

Visual images represent human vision and are not capable of representing some of the most important aspects of a visual scene.
In the case of ammunition scrap, the visual cameras are not able to capture any information from the inside of the ammunition casings.
However, there might be energetics still remaining inside of the scrap pieces even if they are separated.
Therefore, other visual modalities such as x-ray penetration can help increase the accuracy and reliability of the inspection.
The x-ray images clearly capture the gunpowder inside the scrap pieces, which makes them a beneficial resource in the classification of safe/unsafe samples.

\begin{figure}[!t]
	\begin{center}
		\setcounter{subfigure}{0}
		\subfigure[]{
			\centering
			\includegraphics[width=24mm]{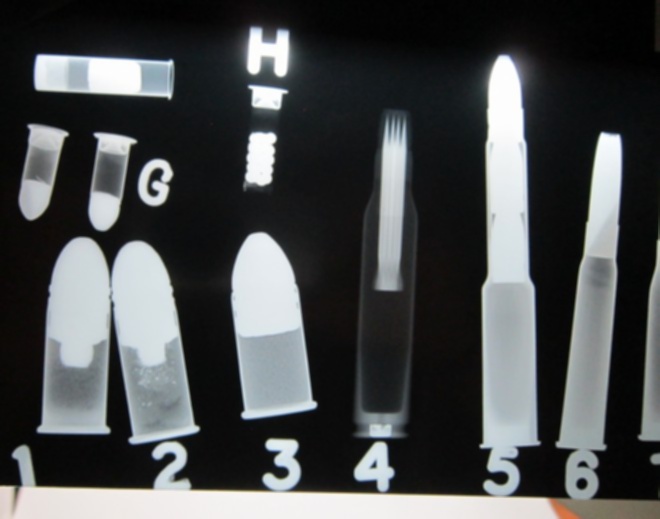}\label{subfig:avg_0}
		} 
		\subfigure[]{
			\centering
			\includegraphics[width=24mm]{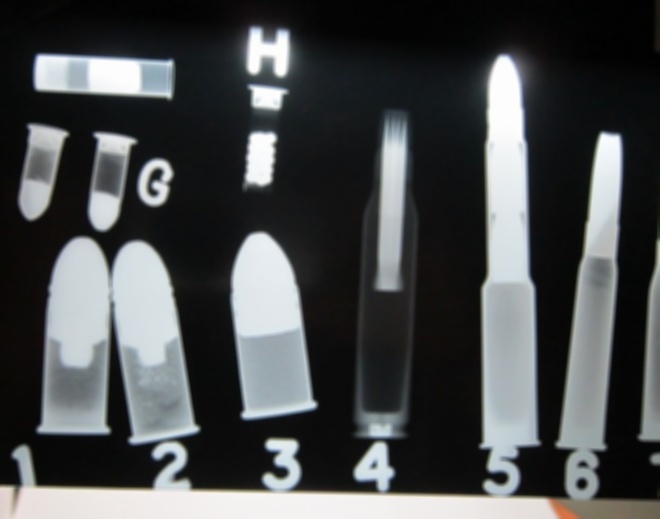}\label{subfig:avg_1}
		} 
		\subfigure[]{
			\centering
			\includegraphics[width=24mm]{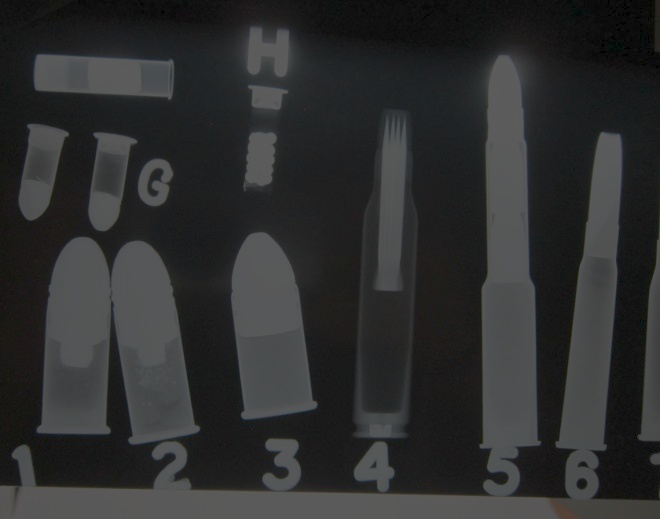}\label{subfig:PL_0}
		} 
		\subfigure[]{
			\centering
			\includegraphics[width=24mm]{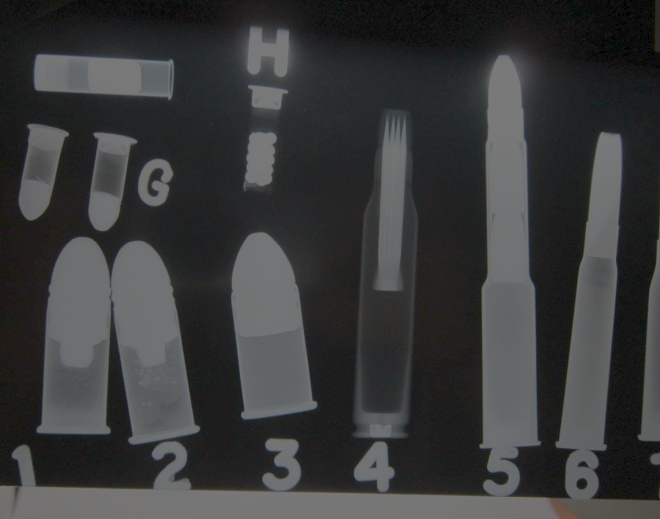}\label{subfig:PL_1}
		} 
		\subfigure[]{
			\centering
			\includegraphics[width=24mm]{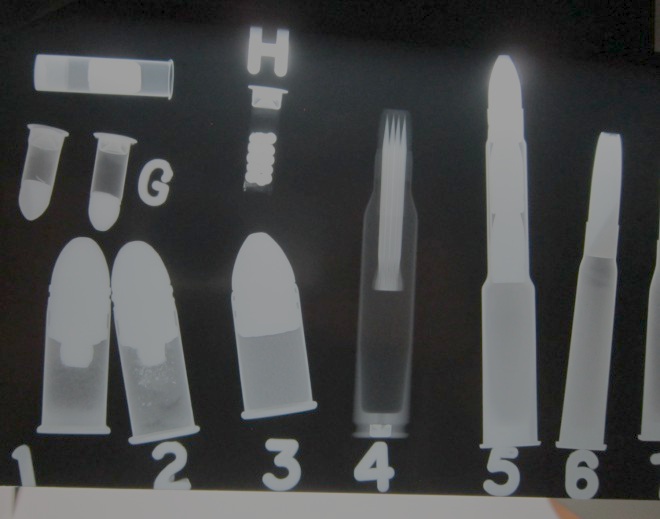}\label{subfig:PL_2}
		} 
		\subfigure[]{
			\centering
			\includegraphics[width=24mm]{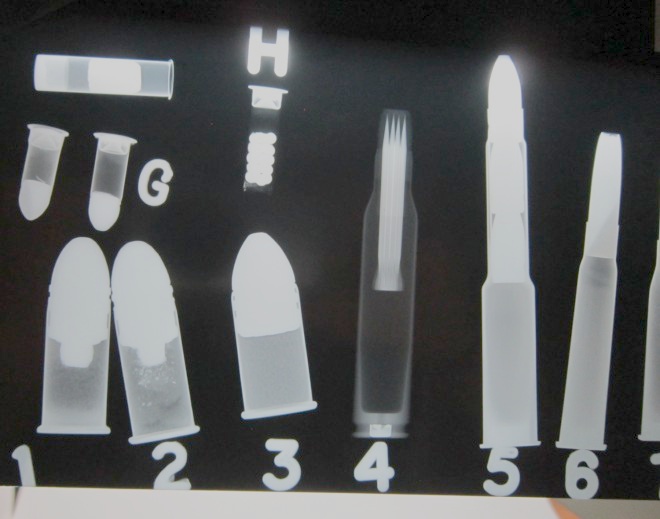}\label{subfig:PL_3}
		}
		\subfigure[]{
			\centering
			\includegraphics[width=24mm]{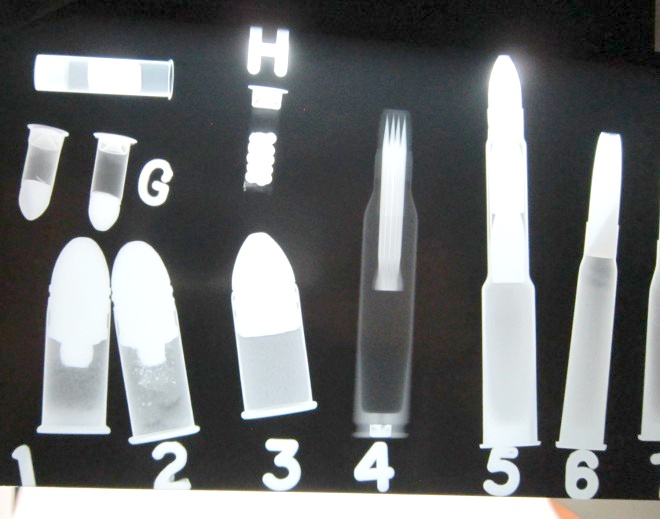}\label{subfig:PL_4}
		} 
		\subfigure[]{
			\centering
			\includegraphics[width=24mm]{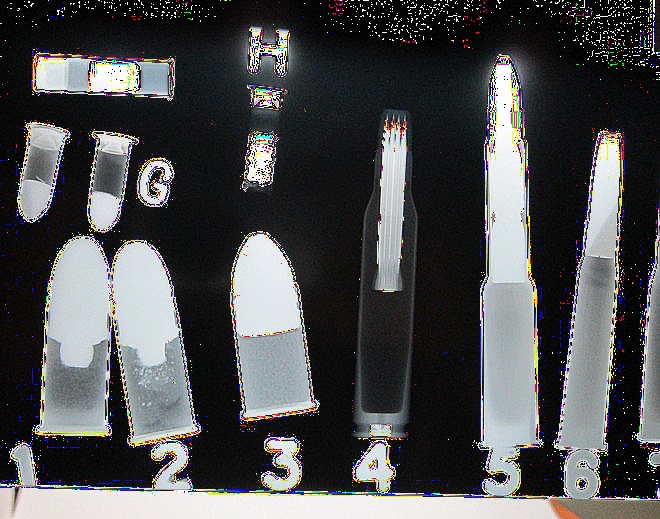}\label{subfig:sharp_1}
		} 
		\caption{The sample images generated using data augmentation. 
			\protect\subref{subfig:avg_0} and \protect\subref{subfig:avg_1} are the results of averaging with kernel sizes 3 and 5, respectively.	
			\protect\subref{subfig:PL_0}, \protect\subref{subfig:PL_1}, \protect\subref{subfig:PL_2}, \protect\subref{subfig:PL_3}, \protect\subref{subfig:PL_4} are the results of power law transformation with gamma values 0.40, 0.45, 0.5, 0.55, and 0.6, respectively,
			and \protect\subref{subfig:sharp_1} is the result of sharpening.
		}
		\label{fig:aug}
	\end{center}
	\vspace*{-6pt} 
\end{figure}

We have gathered a number of x-ray images of ammunition to form a dataset for training the deep CNN model.
Methods based on deep learning require a large number of training data in order to tune the parameters and learn the features.
Since the number of acquired x-ray images is not sufficient to train a deep learning model we have applied a number of data augmentation techniques in order to increase the number of training samples and highlight the features of interest.
We have applied a number of augmentation techniques in the form of spatial transformations, such as histogram equalization, power law, averaging, sharpening, negative, and Gaussian blurring on the input x-ray images.
Image sharpening is applied as the addition of the original image and the high frequency for the purpose of enhancing the edges in images with poor qualities.
The gamma power of image intensities is calculated to compute the power law transformation, which is applied to manipulate the image contrast and perform calibration.
\Cref{fig:aug} illustrates a few examples of data augmentation techniques applied on an x-ray image.

\section{Experiments}  \label{sec_expr}

We conduct a number of experiments using two image datasets to evaluate the performance of YOLOv4 deep learning method in ammunition component detection and classification.
The two datasets are created by collecting visual and x-ray images of ammunition components in addition to full-ammo.
Specifically, we call the first dataset \textbf{Visual Ammunition Component Detection} (VACD), which contains 162 visual images for training along with 50 images for testing.
Each image contains somewhat between one and thirteen scrap pieces where each piece is either a full-ammo or an ammunition component such as casing or projectile.

The second dataset is named \textbf{X-Ray Ammunition Component Detection} (XACD), which is a collection of 108 x-ray images obtained by applying data augmentation techniques on the initial 12 images.
From the 108 x-ray images, 72 are used as training samples and the remaining 36 are utilized as test data.
For each visual or x-ray image, the ground-truth bounding boxes are manually annotated and labeled.
\Cref{table:training} summarizes the three training data sets.

\begin{table}
\setlength{\tabcolsep}{0.5em} 
\renewcommand{\arraystretch}{1.4}
	\centering
\caption{The Ammunition Component Datasets}
\scalebox{.9}{
\begin{tabular}{c c c c}
	\hline
	Dataset & Imaging modality & Training samples  & Testing samples\\
	\hline
	\hline
	VACD & Visible & 162 & 50 \\
	XACD & X-Ray & 72 & 36  \\
	\hline
\end{tabular}}
\label{table:training}
\end{table}


For evaluating the quantitative results of the object detection method, three performance measures are utilized as follows:
\begin{equation} \label{eq:evaluate_metrics}
	\begin{cases}
		PRE=T_P/(T_P+F_P)\\
		REC=T_P/(T_P+F_N)\\
		F_1= 2 \times (PRE \times REC)/(PRE+REC)
	\end{cases}
\end{equation}
where $T_P$, $F_P$, and $F_N$ are the true positive, false positive, and false negative instances, respectively.
$PRE$, $REC$, and $F_1$ refer to precision, recall, and F1-score, respectively.

\begin{figure}[!t]
	\begin{center}
		\subfigure{
			\centering
			\includegraphics[width=24mm]{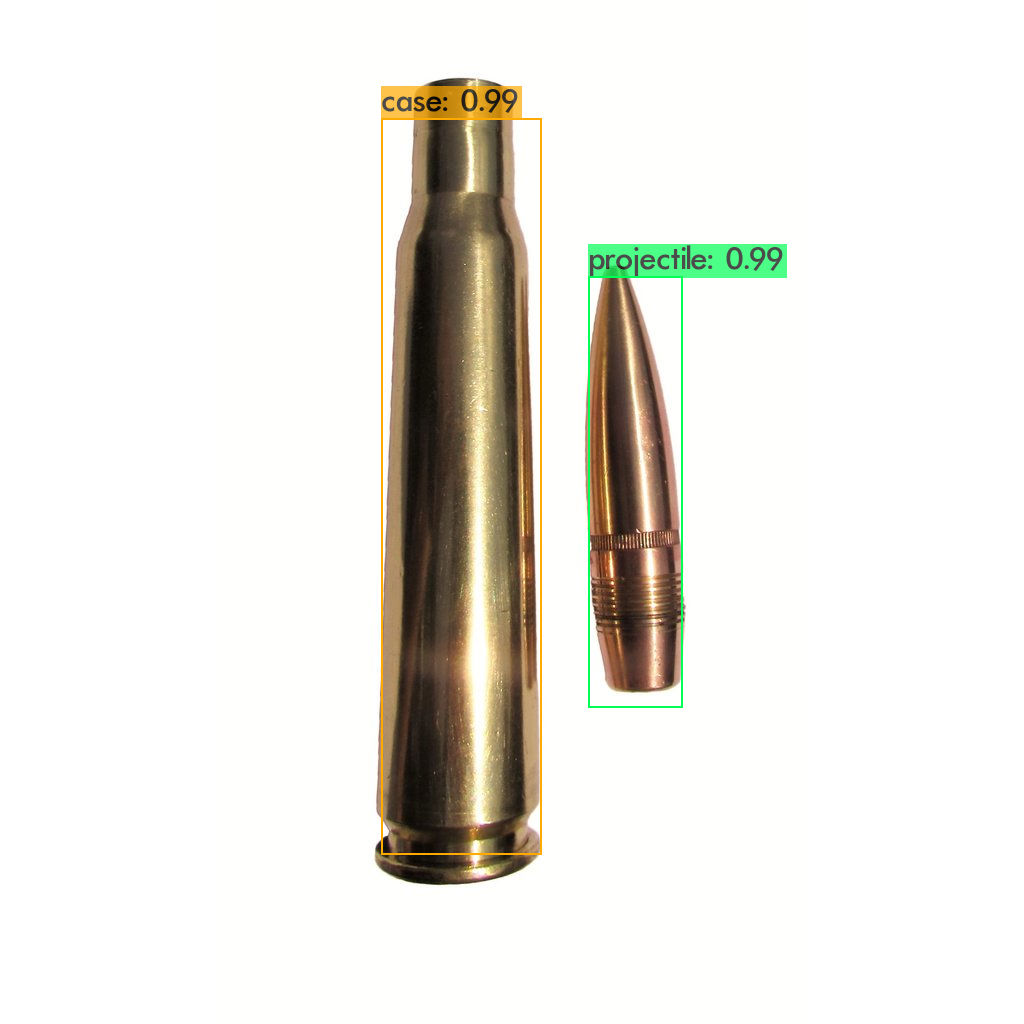}
		} 
		\subfigure{
			\centering
			\includegraphics[width=24mm]{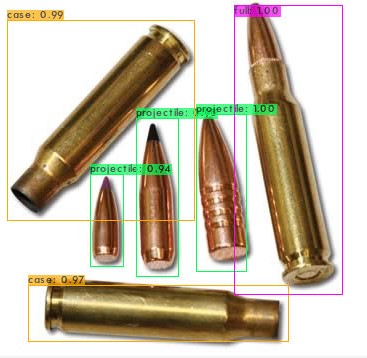}
		} 
		\subfigure{
			\centering
			\includegraphics[width=24mm]{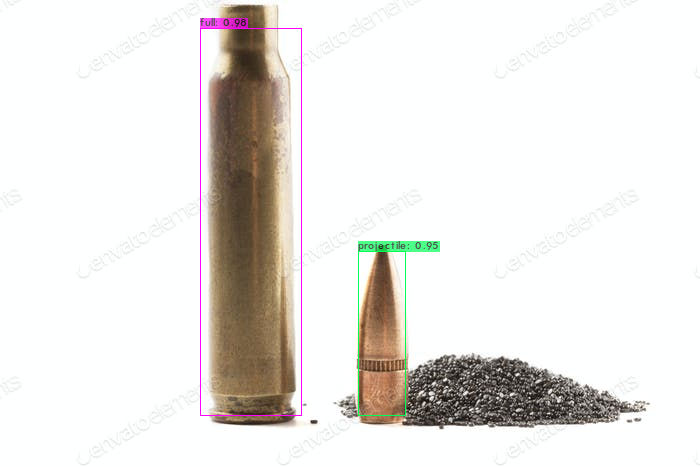}
		} 
		\subfigure{
			\centering
			\includegraphics[width=24mm]{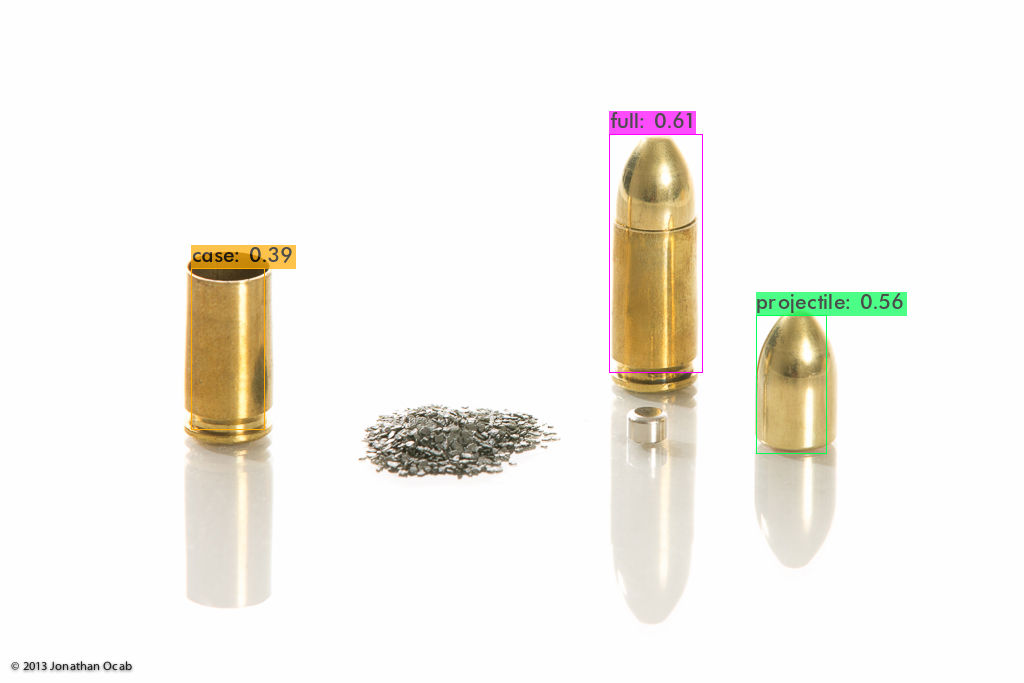}
		} 
		\subfigure{
			\centering
			\includegraphics[width=24mm]{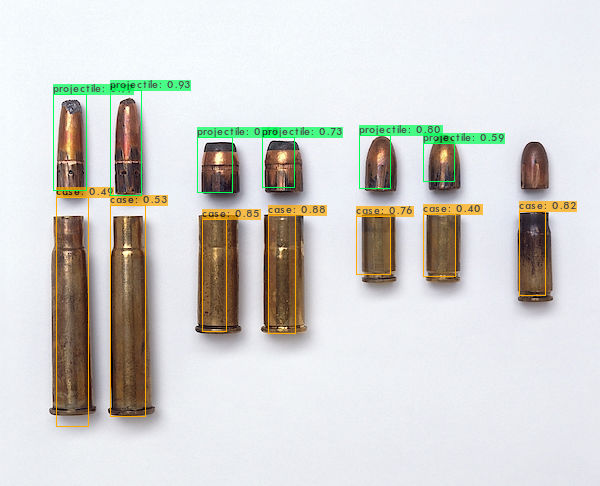}
		} 
		\subfigure{
			\centering
			\includegraphics[width=24mm]{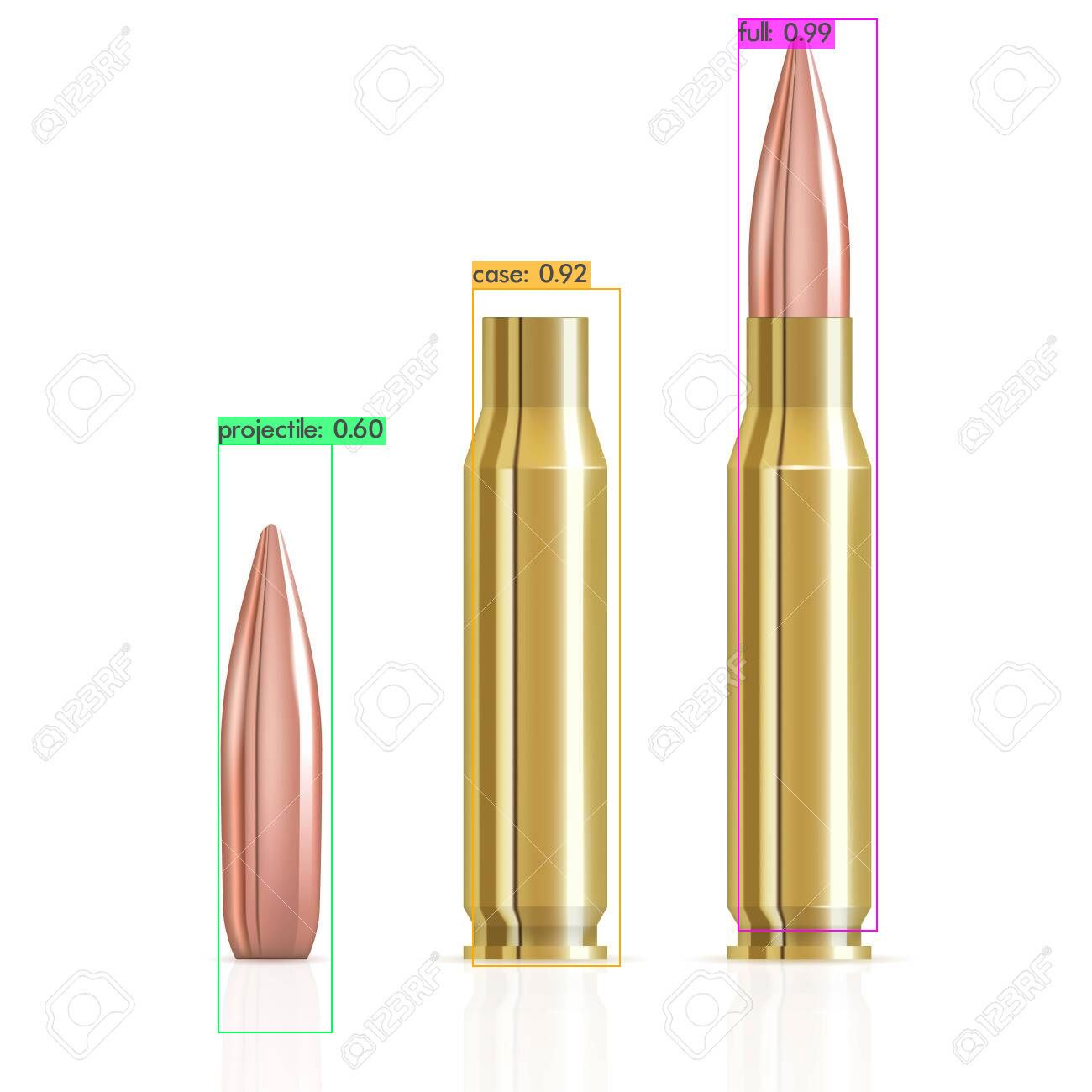}
		} 
		\subfigure{
			\centering
			\includegraphics[width=24mm]{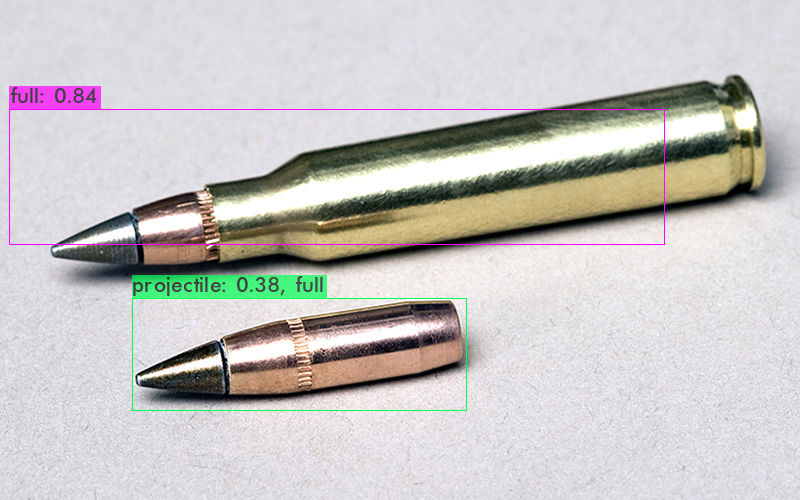}
		} 
		\subfigure{
			\centering
			\includegraphics[width=24mm]{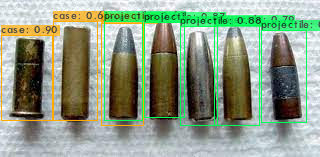}
		}  
		\caption{Sample detection and classification results of applying tiny-YOLOv4 on the test images of the VACD dataset.
		}
		\label{fig:vacd}
	\end{center}
	\vspace*{-6pt} 
\end{figure}

We used a public github repository \cite{alexey_2021_5622675} for the experiments.
For parameter settings, the batch sizeis 64, learning rate is 0.001, momentum is 0.973, and decay is selected to be 0.0005.
The tiny-YOLOv4 backbone is a CSP that contains CBL, cross-stage, and residual features along with skip connection layers.
It applies two detectors at the end head.
More details can be found at the repository \cite{alexey_2021_5622675}.
The first experiments were carried out using the VACD dataset for training a tiny-YOLOv4 model.
\Cref{tab:tiny_vacd} shows the results of the experiments conducted on the VACD data in terms of the performance measures.
The confidence threshold is set to $0.25$ and the intersection over union (IoU) threshold is computed as $54.63\%$.
The mean average precision for this dataset reached $84.94\%$ and the total detection time was $6$ seconds.
\Cref{fig:vacd} shows a number of detection results of testing the tiny-YOLOv4 model against the images that are unseen during the training.

\begin{table}[!h]
	\setlength{\tabcolsep}{0.5em} 
	\renewcommand{\arraystretch}{1.4}
	\caption{The quantitative results of testing tiny-YOLOv4 on the VACD data}
	\centering
	\begin{tabular}{c | c c c c c c}
		\hline
		&  TP  & FP & FN & Precision & Recall & F1-score\\	
		\hline
		\hline
		Full-ammo & 159 & 50 & 21 & 76.08\% & 88.33\% & 81.75\% \\
		Casing & 92 & 104 & 10 & 46.94\% & 90.2\% & 61.74\% \\
		Projectile & 84 & 25 & 9 & 77.06\% & 90.32\% & 83.17\% \\
		\hline
		Total & 335 & 179 & 40 & 65\% & 89\% & 75\% \\
		\hline
	\end{tabular}
	
	\label{tab:tiny_vacd} 
\end{table}

\begin{figure}[!t]
	\begin{center}
		\subfigure{
			\centering
			\includegraphics[width=24mm]{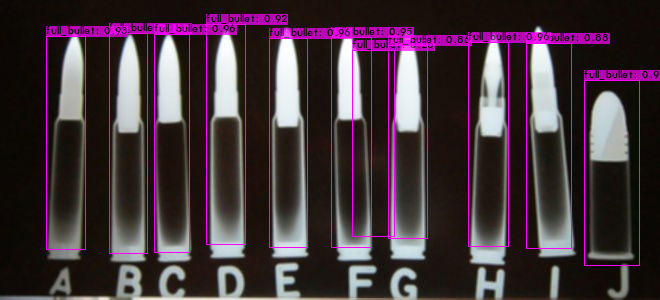}
		} 
		\subfigure{
			\centering
			\includegraphics[width=24mm]{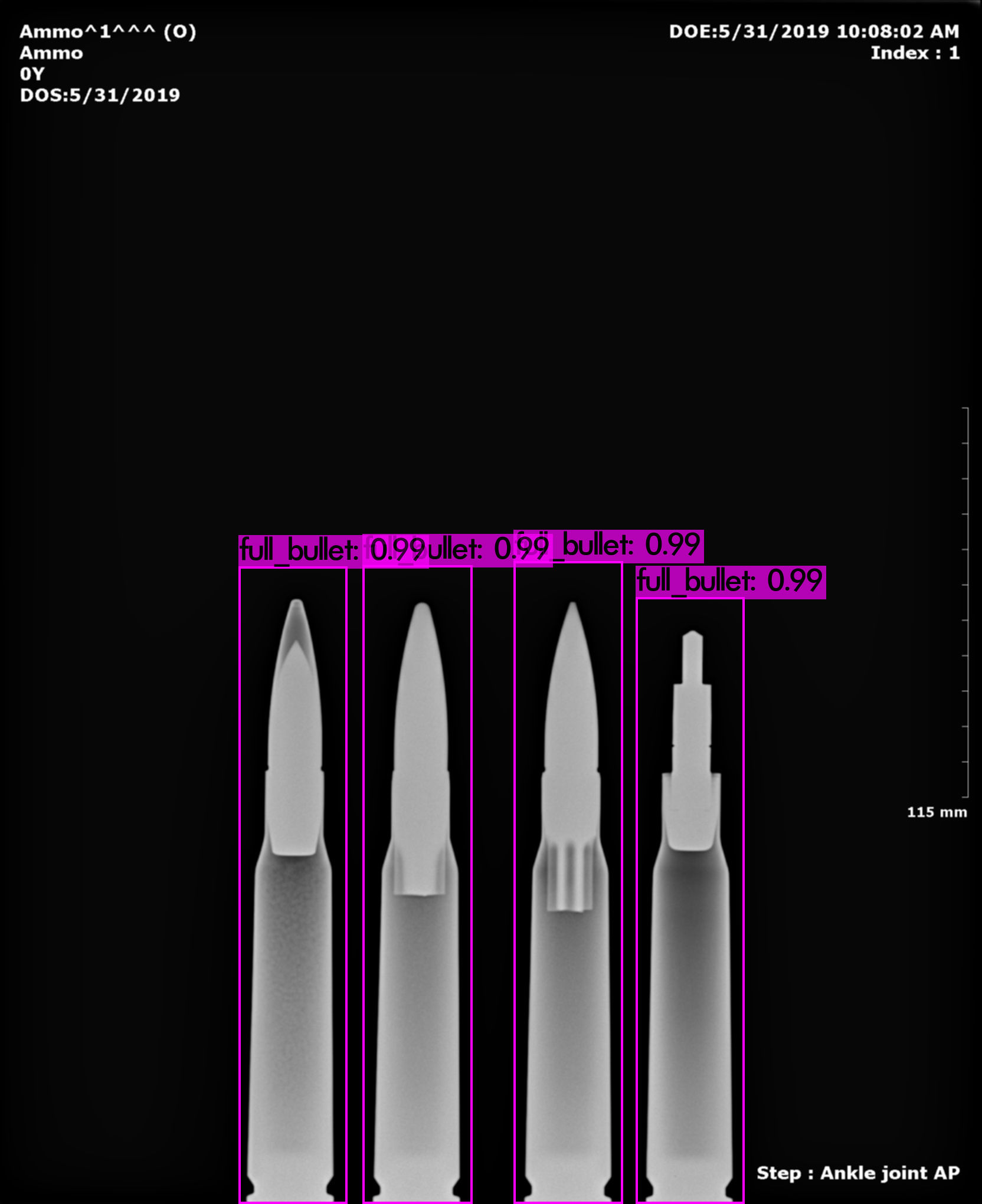}
		} 
		\subfigure{
			\centering
			\includegraphics[width=24mm]{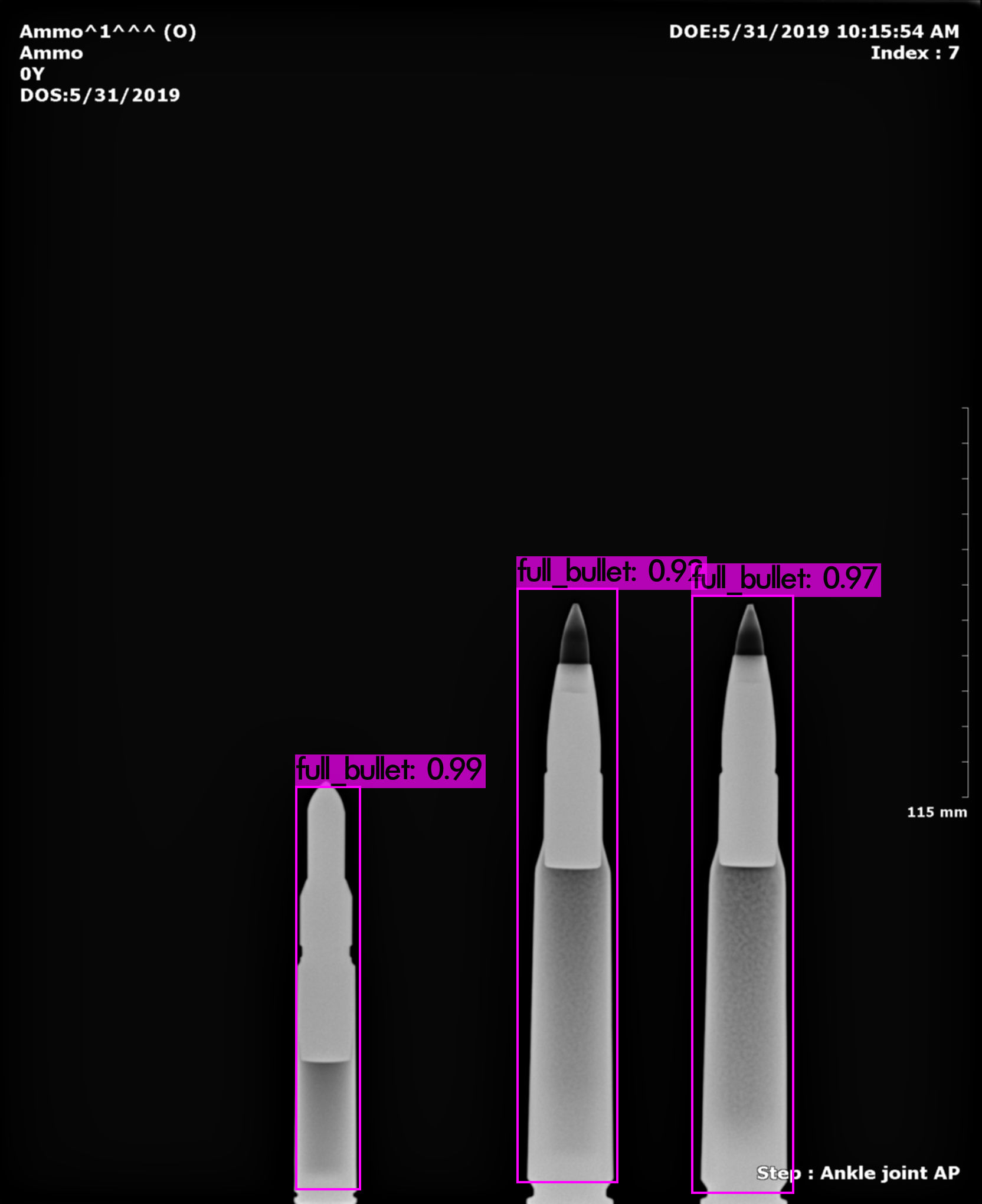}
		} 
		\subfigure{
			\centering
			\includegraphics[width=24mm]{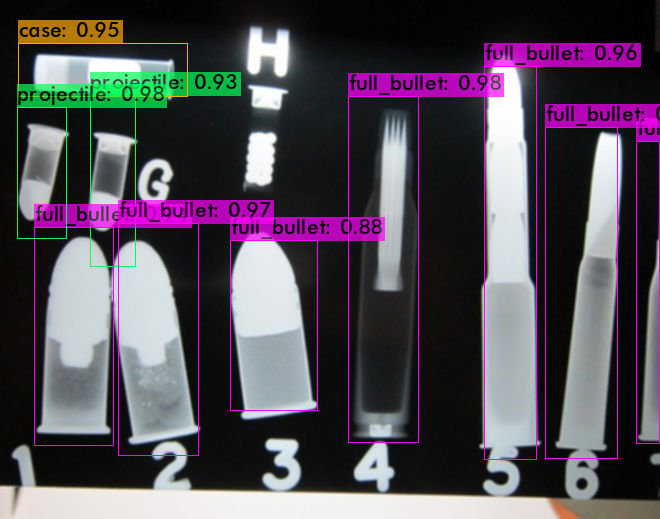}
		} 
		\subfigure{
			\centering
			\includegraphics[width=24mm]{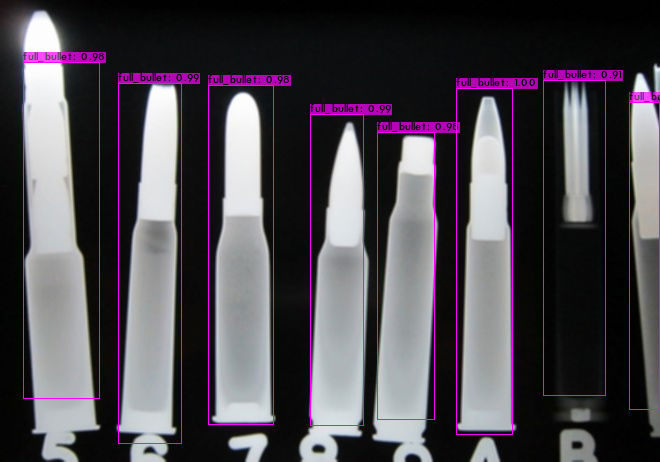}
		} 
		\subfigure{
			\centering
			\includegraphics[width=24mm]{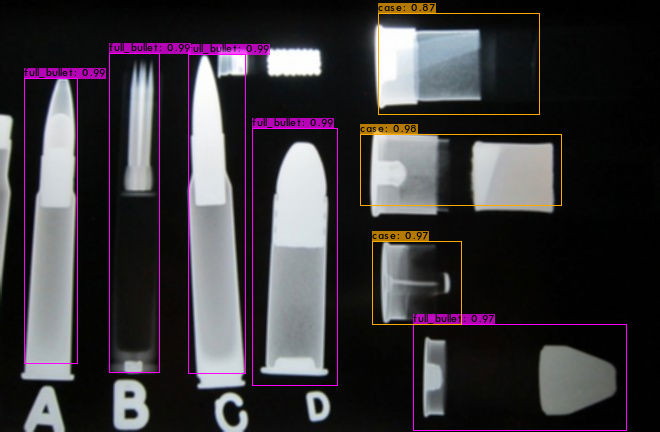}
		} 
		\subfigure{
			\centering
			\includegraphics[width=24mm]{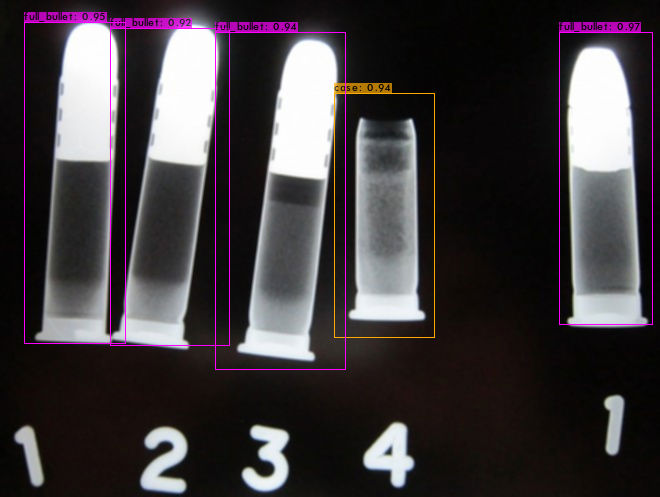}
		} 
		\subfigure{
			\centering
			\includegraphics[width=24mm]{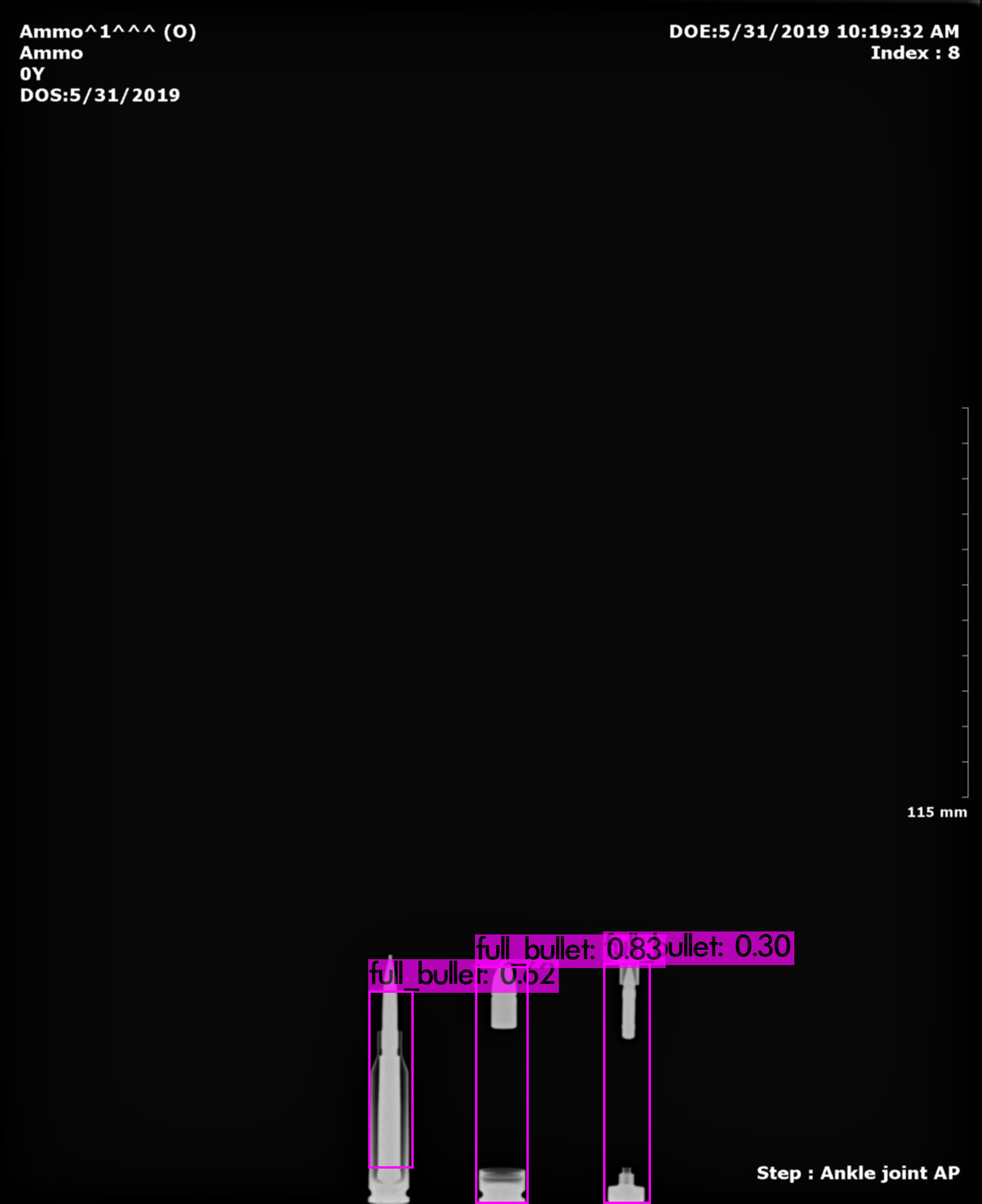}
		}  
		\caption{Sample detection and classification results of applying tiny-YOLOv4 on the test images of the XACD dataset.
		}
		\label{fig:xacd}
	\end{center}
	\vspace*{-6pt} 
\end{figure}

The second set of experiments was carried out using the XACD dataset for training a tiny-YOLOv4 model.
\Cref{tab:tiny_xacd} shows the results of the experiments conducted on the XACD data in terms of the performance measures.
The confidence threshold is set to $0.25$ and the intersection over union (IoU) threshold is computed as $50\%$.
The mean average precision for this dataset reached $99.92\%$ and the total detection time was $8$ seconds.
\Cref{fig:xacd} illustrates sample detection results of testing the tiny-YOLOv4 model against the images that are unseen during the training.

\begin{table}[!h]
	\setlength{\tabcolsep}{0.5em} 
	\renewcommand{\arraystretch}{1.4}
	\caption{The quantitative results of testing tiny-YOLOv4 on the XACD data}
	\centering
	\begin{tabular}{c | c c c c c c}
		\hline
		&  TP  & FP & FN & Precision & Recall & F1-score\\	
		\hline
		\hline
		Full-ammo & 499 & 21 & 1 & 95.96\% & 99.8\% & 97.84\% \\
		Casing & 41 & 6 & 7 & 87.23\% & 85.42\% & 86.32\% \\
		Projectile & 18 & 0 & 4 & 100\% & 81.82\% & 90\% \\
		\hline
		Total & 558 & 27 & 12 & 95\% & 98\% & 97\% \\
		\hline
	\end{tabular}
	
	\label{tab:tiny_xacd} 
\end{table}

\begin{figure} [!h]
	\centering
	\includegraphics[width=1.0\linewidth]{"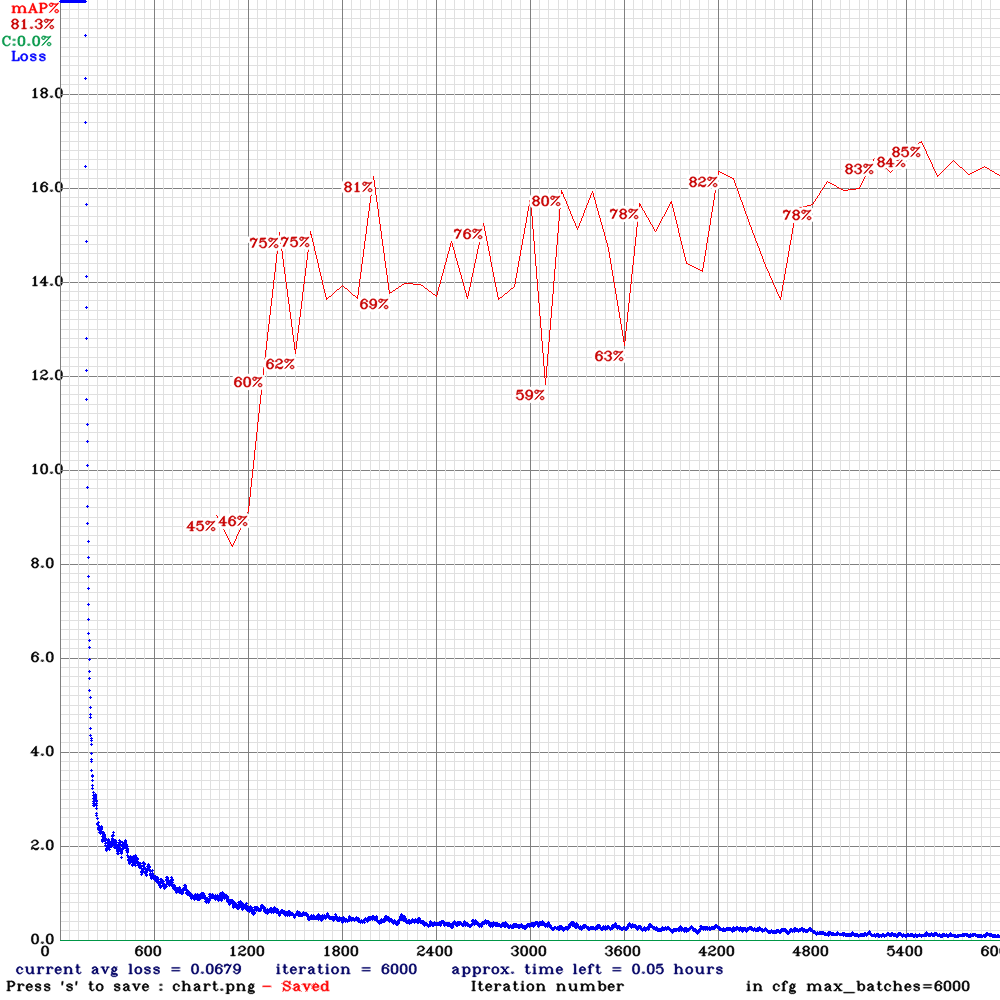"}
	\caption{changes in the loss and mAP in 6000 iterations of training tiny-YOLOv4 against the VACD data.}
	\label{fig:map}
\end{figure}

The hardware specification used for the experiments is a 3.4 GHz processor, 16 GB RAM, and an Nvidia GTX-745 graphics processing unit (GPU).
The time spent on the training of the tiny-YOLOv4 model using the VACD dataset for 6000 iterations was approximately four hours.
\Cref{fig:map} illustrates the growth of mean average precision (mAP) in terms of training iterations using the VACD data.

Our findings in this study demonstrate the need of deep learning models for a considerable amount of data and computational resources in comparison with the traditional methods.
Additionally, however necessary for supervised learning, the manual labeling and annotating the collected data is a tedious and time-consuming process.
The extracted features are not easily interpretable by human experts and the complexity of numerous hidden layers in the deep network architectures increases as the model uses more parameters.
Another drawback of the deep learning methods is their low ability of generalization.

Despite the limitations, when tested on data similar to the training samples, a well-trained deep model demonstrates high performance in the task that it is trained for.
All the YOLOv4 models trained using the collected datasets showed great accuracy in detecting ammunition components and full-ammo when tested against visible or x-ray images.
The YOLOv4 method proved to be fast enough for use in applications with real-time requirements.
Visual and x-ray imaging systems can be deployed on the inspection site.
The trained models can be applied in real-time during the ammunition metallic scrap inspection in order to discriminate the Material Documented as Safe (MDAS), which involves the projectile and casing samples, and Material Potentially Possessing Energetic Hazard (MPPEH) which refers to the full-ammo class.
The potentially dangerous scrap pieces should be separated for further inspections.

\section{Conclusions}  \label{sec_con}
The representative YOLOv4 object detection method is applied for the task of ammunition component detection and classification in images of the visible and the x-ray range.
The purpose of this classification is to aid with the automatic inspection of ammunition scrap in the process of reducing dangerous properties prior to recycling.
Two image databases are gathered and annotated for the task of object detection.
First dataset, Visual Ammunition Component Detection (VACD), is a set of visible images of ammunition components and full-ammo pieces.
The second database, X-Ray Ammunition Component Detection (XACD), is a set of x-ray images of ammunition components, which is enhanced by several data augmentation techniques.
The x-ray images are able to illustrate the insides of the ammunition casings, which helps indicate whether the scrap piece contains any energetics or not.
As a general rule of thumb, we consider full-ammo to be explosive hazard due to the possibility of remaining energetics inside.
The potentially dangerous scrap pieces should be separated and go through the incineration process again before being moved to the recycling unit.
The experimental evaluations using the collected datasets demonstrate the feasibility of the YOLOv4 method in object detection and classification in real-time applications.

\bibliographystyle{splncs04}
\bibliography{mldmcv}

\end{document}